\documentclass[letterpaper, 10 pt, conference]{ieeeconf}
\IEEEoverridecommandlockouts
\usepackage{amsmath,amssymb}
\usepackage{booktabs}
\usepackage{graphicx}
\usepackage{multirow}
\usepackage{xcolor}
\usepackage{url}
\usepackage{microtype}
\microtypesetup{protrusion=false}
\usepackage{ragged2e}
\usepackage{placeins}
\usepackage[hidelinks]{hyperref}
\usepackage{latexml}

\iflatexml
\newcommand{\tabletitle}[1]{\caption{#1}}
\else
\newcommand{\tabletitle}[1]{%
  \refstepcounter{table}%
  \par\begingroup\setlength{\parskip}{0pt}\small\justifying\noindent
  \textbf{TABLE~\Roman{table}:}~#1\par\endgroup\vspace{1.5pt}}
\fi

\newcommand{\method}{CoRE}

\newcommand{\srfrac}[1]{#1{\tiny/100}}
\newcommand{\speedratio}[1]{\iflatexml\text{#1\texttimes}\else#1\times\fi}

\title{\LARGE \bf
\method: Learning Collaboration-Role Experts for\\
Decentralized Collaborative Manipulation with One Policy}

\author{Yanan Zhou$^{1}$, Zhaoyan Qian$^{1}$, Zihao Li$^{1}$, Mingyuan Ba$^{1}$, Ranpeng Qiu$^{2}$, and Weiming Zhi$^{1*}$%
\thanks{$^{1}$School of Computer Science, The University of Sydney, Australia.}%
\thanks{$^{2}$R. Qiu is with the College of Computer Science and Technology, Zhejiang University of Technology, China.}%
\thanks{$^{*}$Corresponding author: \texttt{Weiming.Zhi@sydney.edu.au}.}}

\iflatexml\else
\makeatletter
\IEEEaftertitletext{%
    \begin{minipage}{\textwidth}
        \centering
        \includegraphics[width=\textwidth]{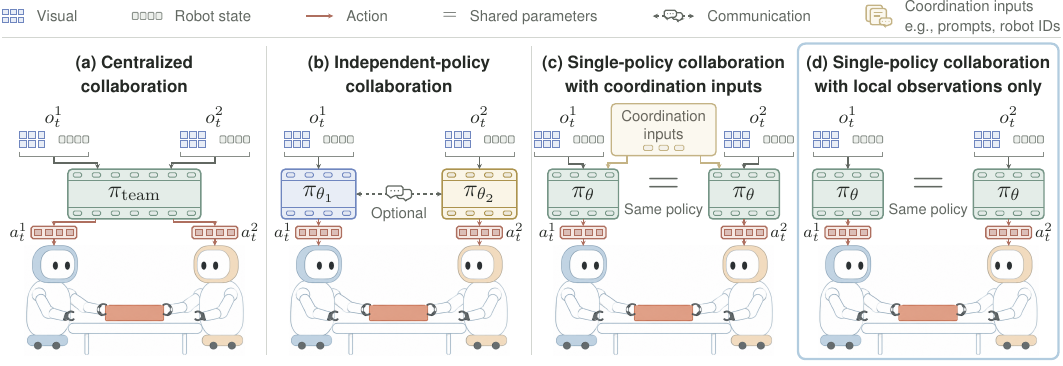}
        \def\@captype{figure}%
        \caption{\textbf{Multi-robot collaboration paradigms.} $o_t^i$ and $a_t^i$ are robot $i$'s local observation and action at time $t$; $\pi$ and $\theta$ denote a policy and its parameters. (b) allows communication; (c) adds coordination inputs. The blue frame marks single-policy decentralized collaboration from local observations only.}
        \label{fig:information-boundary}
    \end{minipage}\vspace{8pt}%
}
\makeatother
\fi

\begin{document}
% Apply IEEE bibliography controls without creating a numbered reference.
\makeatletter
\immediate\write\@auxout{\string\citation{CoREBSTcontrol}}
\makeatother
% Keep natural vertical spacing instead of stretching headings to align column bottoms.
\flushbottom
\maketitle
\iflatexml
\begin{figure*}
\centering
\includegraphics[width=\textwidth]{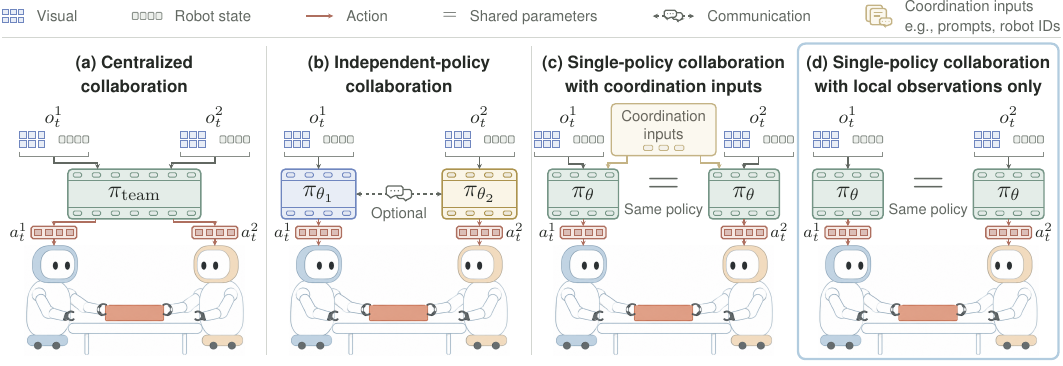}
\caption{\textbf{Multi-robot collaboration paradigms.} $o_t^i$ and $a_t^i$ are robot $i$'s local observation and action at time $t$; $\pi$ and $\theta$ denote a policy and its parameters. (b) allows communication; (c) adds coordination inputs. The blue frame marks single-policy decentralized collaboration from local observations only.}
\label{fig:information-boundary}
\end{figure*}
\fi
\thispagestyle{empty}
\pagestyle{empty}

\begin{abstract}
Collaborative manipulation requires robots to perform complementary actions as interactions unfold. We study single-policy decentralized collaboration: every robot runs the same policy from its visual observations and proprioception, without task prompts, identity labels, or inter-robot messages. The challenge is to learn complementary team behaviors within shared parameters and select appropriate actions from each robot's local observations. We introduce \method{}, which learns \underline{Co}llaboration-\underline{R}ole \underline{E}xperts from pooled multi-task, multi-robot demonstrations. Fused appearance and geometry provide local interaction evidence. Query-conditioned cross-attention experts provide adaptable prediction paths, which a local router combines at each action-chunk position. During training, an action--expert alignment loss supervises expert selection using relative forced-route prediction errors against demonstrations under fixed inputs, without role labels. Across simulation benchmarks, \method{} achieves the highest average performance among evaluated decentralized methods. Physical experiments demonstrate effective collaboration across diverse manipulation tasks and robustness to partner delays and slowdowns. Project~page: \url{https://aus.bot/research/core/}.
\end{abstract}

\section{Introduction}

Imitation learning of visuomotor policies is enabling individual robots to acquire increasingly diverse manipulation skills. Yet many real-world tasks require multiple robots, such as moving large furniture or performing collaborative assembly. A common approach is centralized control, which aggregates observations from all robots in a single controller and generates actions for the entire team. Everyday human collaboration suggests another possibility. People often coordinate without discussing each action, using visible changes in the object and their partners' behavior to adjust their movements and timing. This motivates learning decentralized collaboration from demonstrations.

In Fig.~\ref{fig:information-boundary}, panel~(a) centralizes team observations and action prediction, whereas panel~(b) assigns a separate policy to each robot. Sequential Asymmetric Imitation (SAI)~\cite{chen2026sai} learns coupled policies that collaborate from local observations without communication. Panels~(c) and~(d) instead share one policy across robots. As tasks or team composition change, parameter sharing provides a single model to update and deploy throughout the team. Panel~(c) additionally uses explicit coordination inputs, such as exchanged capability descriptors~\cite{howell2023capabilities}. We study \emph{single-policy decentralized collaboration} in panel~(d): each robot predicts its own actions from visual observations and proprioception, without task prompts, robot identities, role labels, or inter-robot messages.

This setting requires a shared policy to produce complementary behaviors as the interaction unfolds. During an object handover, the same robot may first maintain its grasp, then offer the object to a partner, and finally release it. The appropriate action depends on the progress of the interaction rather than robot identity alone. We refer to this interaction-dependent action responsibility as a \emph{collaboration role}. Learning from pooled multi-task, multi-robot demonstrations requires both preserving the behaviors needed for different responsibilities and selecting appropriate actions from local observations. Demonstrations specify actions, but not how to organize and select these behavioral capabilities.

To address this challenge, we introduce Collaboration-Role Experts (\method{}). Cross-attention experts provide separately adaptable paths for complementary behaviors within a shared action-chunking policy, while a local router selects their contributions from appearance--geometry evidence and action queries. During training, action--expert alignment compares forced-route predictions from the full decoder against demonstrations under fixed inputs. Their relative errors guide expert selection alongside imitation, without role labels.

CoRE achieves the highest average performance among evaluated decentralized methods across three simulation benchmarks, including 84.2\% success on RoboFactory~\cite{qin2025robofactory}. Real-world experiments achieve 91.3\% success across five tasks spanning coupled motion, asymmetric cooperation, and sequential placement. Sharing comparisons and ablations examine team-demonstration sharing and expert selection. Routing visualizations show expert reuse across robots and changes across stages; partner perturbations test robustness to delays and slowdowns.

Our contributions are as follows.

\iflatexml\begin{enumerate}\else
\begin{enumerate}[\setlength{\labelindent}{0pt}\setlength{\labelsep}{0.45em}]
\fi
    \item We formulate single-policy decentralized collaboration as a visuomotor imitation learning problem. Multi-task, multi-robot demonstrations teach complementary team behaviors to one policy, which each robot runs independently to predict collaborative actions from local observations.
    \item We introduce \method{}, which organizes complementary behaviors through cross-attention experts and uses action--expert alignment to supervise local routing. Demonstrations thereby guide both action learning and expert selection for the current collaboration requirements.
    \item We validate the performance advantages of one shared policy through simulation benchmarks and physical experiments, and demonstrate robustness to partner pauses and slowdowns across coupled, asymmetric, and sequential collaboration.
\end{enumerate}

\begin{figure*}[t]
    \vspace*{5pt}% Keep top-float ink inside the 0.75-inch margin.
    \centering
    \includegraphics[width=\textwidth]{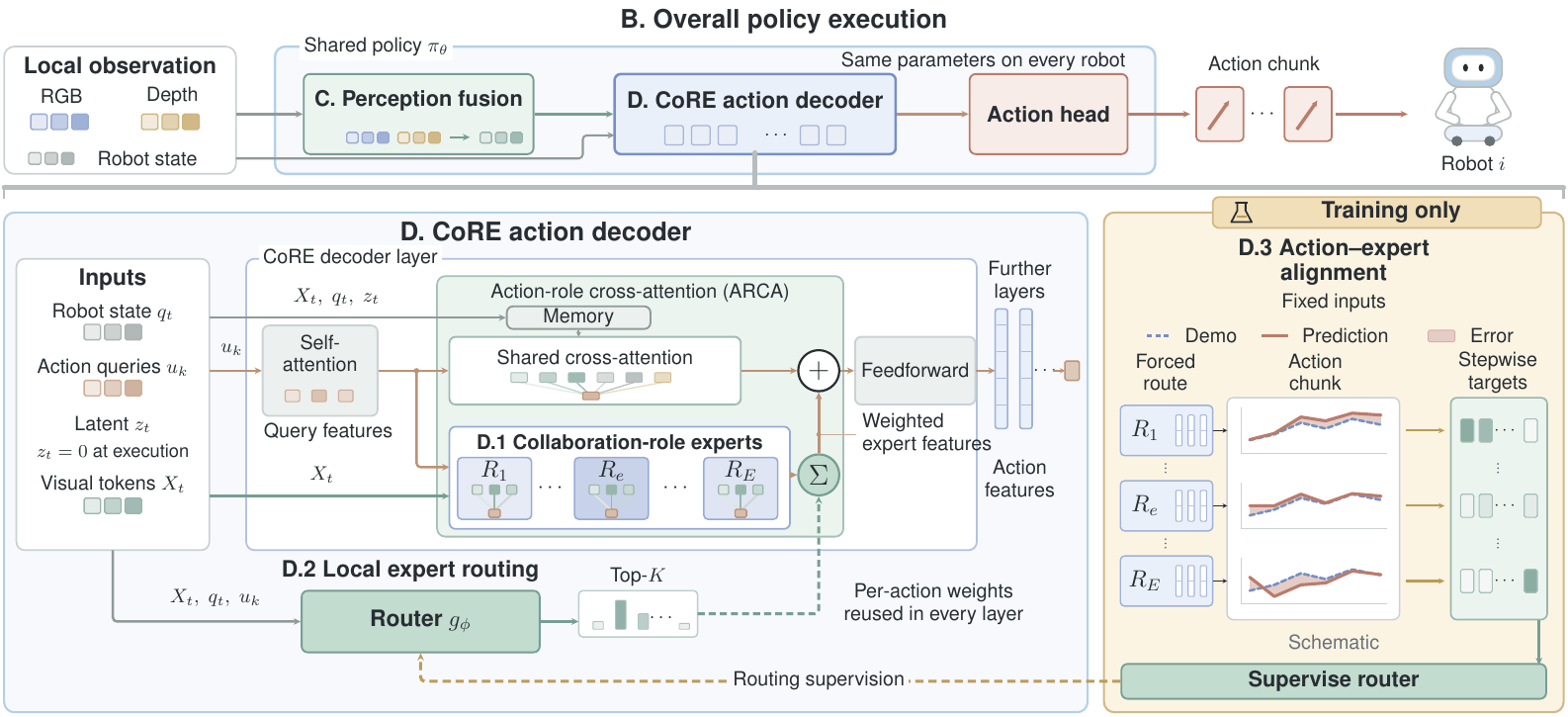}
    \caption{\textbf{CoRE overview.} A shared policy predicts action chunks from local observations. Collaboration-role experts augment shared cross-attention, with stepwise routing supervised by action--expert alignment. Shading illustrates schematic action error.}
    \label{fig:stereocore-overview}
    \vspace{-5pt}% Recover the inset from the space below the float.
\end{figure*}

 \section{Related Work}

\noindent\textbf{Visuomotor policy learning.}
Regression-based visuomotor policies predict action chunks for bimanual manipulation~\cite{zhao2023act}; Diffusion Policy models multimodal action sequences~\cite{chi2023diffusionpolicy}; and $\pi_0$ combines pretrained vision--language representations with continuous action generation across robot embodiments~\cite{black2024pi0}. StereoPatch strengthens spatial perception through appearance--geometry fusion~\cite{zhou2026stereopatch}. At deployment, PATCH monitors action-conditioned local visual changes to trigger intervention and policy resumption~\cite{zhou2026patch}. These advances provide foundations for action prediction, spatial perception, and execution monitoring. CoRE studies how to learn complementary team behaviors within one policy and select actions from each robot's local observations.

\noindent\textbf{Collaborative manipulation.}
Cooperative impedance control regulates object motion and internal forces~\cite{caccavale2008impedance}, while local force feedback can support collaboration without explicit communication~\cite{tsiamis2015implicit}. TriManPolicy retimes demonstrations to train a centralized visuomotor policy that jointly controls three arms~\cite{zhao2026trimanpolicy}. Learned collaboration also uses Gaussian scene representations in GauDP~\cite{wang2025gaudp}, partner-behavior embeddings in LatentToM~\cite{he2025latenttom}, and collaborative latents conditioned on local images and task instructions in CLS-DP~\cite{park2026clsdp}. MIMIC-D jointly trains distinct diffusion policies for decentralized coordination~\cite{dong2025mimicd}; SAI trains separate coupled policies through a sequential partner-training curriculum~\cite{chen2026sai}. CHORUS executes copies of one vision--language--action policy without inter-robot communication, using robot-identifying prompts~\cite{doshi2026chorus}. CoRE learns complementary team behaviors through shared collaboration-role experts, supervising their selection with demonstrated-action errors. Each robot selects expert contributions based on its local observations.

\noindent\textbf{Mixture-of-experts learning.}
Mixture-of-experts (MoE) models allocate capacity through input-dependent expert combinations~\cite{jacobs1991adaptive}. Sparse gating~\cite{shazeer2017moe,fedus2022switch} and expert-choice routing~\cite{zhou2022expertchoice} scale conditional computation by selecting which inputs and experts interact. Sparse Diffusion Policy applies expert modularity to robotic action prediction~\cite{wang2025sparsedp}. The central issue for CoRE is how to select experts for the current collaborative action. Existing routing guidance includes pretrained observation--action latents in LAR-MoE~\cite{rodriguez2026larmoe}, semantic concept priors in CoRDE~\cite{huang2026corde}, and vision-teacher targets in TGR-MoE~\cite{kada2026tgr}. Expert--router coupling uses proxy-token activations to align routing with expert capabilities~\cite{lv2026erc}; alternative routes have also been evaluated through next-token predictions~\cite{yoon2026misrouted}. CoRE compares the shared decoder's forced-route predictions against demonstrations under fixed inputs. Their relative errors supervise each action query's routing during imitation learning, linking expert selection to complementary action requirements without a separate teacher or role labels.
 \section{CoRE: Collaboration-Role Experts}
\label{sec:core-method}
\begingroup
\setlength{\abovedisplayskip}{1pt plus 1pt minus 0.5pt}
\setlength{\belowdisplayskip}{1pt plus 1pt minus 0.5pt}
\setlength{\belowdisplayshortskip}{1pt plus 1pt minus 0.5pt}
\setlength{\abovedisplayshortskip}{1pt plus 1pt minus 0.5pt}

\subsection{Problem Formulation}

We study single-policy decentralized collaboration under centralized training with decentralized execution (CTDE)~\cite{lowe2017maddpg}. In each task $m\in\mathcal T$, $N_m$ robots coordinate their actions and timing to complete a common manipulation goal. Demonstrations capture the complementary behaviors required from each robot during these interactions.

At time $t$, robot $i$ receives a local observation $o_t^i$ comprising its available camera views (RGB and registered depth) and its proprioceptive state $q_t^i$. Its demonstrated action chunk $A_t^i$ contains $H$ consecutive actions. The team's demonstrated behavior is represented by $\mathbf A_t^m=(A_t^1,\ldots,A_t^{N_m})$. Training uses local observation--action pairs $(o_t^i,A_t^i)$ from each dataset $\mathcal D_{m,i}$, pooled as
\begin{equation}
\mathcal D=
\bigcup_{m\in\mathcal T}
\bigcup_{i=1}^{N_m}
\mathcal D_{m,i}.
\end{equation}

We learn a single policy $\pi_\theta$, with each robot independently predicting $\hat A_t^i=\pi_\theta(o_t^i)$. These local predictions aim to reproduce the demonstrated team behavior:
\begin{equation}
\bigl(
\pi_\theta(o_t^1),\ldots,\pi_\theta(o_t^{N_m})
\bigr)
\approx
\mathbf A_t^m.
\end{equation}
This joint notation describes the team's behavior; each policy copy receives only its own local observation. All tasks and robots share the same parameters $\theta$, while task and robot indices identify data sources and are not policy inputs. Execution uses no task prompts, robot identities, role labels, or inter-robot messages.

\subsection{Overview}

Figure~\ref{fig:stereocore-overview} outlines CoRE. Fused observations pass through an action decoder and action head to predict action chunks. Our core design combines separately trainable collaboration-role experts with a local router that allocates their contributions at each action position. Action--expert alignment compares predictions from alternative expert routes with demonstrated actions to supervise expert allocation, complementing the imitation loss on the selected expert mixture at each position.

\subsection{Local Spatial Evidence for Collaboration}

We assume each robot's local observations contain sufficient evidence for its demonstrated collaborative actions. Spatial distinctions such as gripper--object distances help distinguish similar appearances requiring different motions. We adopt StereoPatch~\cite{zhou2026stereopatch} to fuse RGB appearance features from pretrained DINOv3~\cite{simeoni2025dinov3} and registered-depth features from pretrained DeFM~\cite{patel2026defm} on a shared image-patch grid, preserving spatial correspondence between appearance and geometry. The resulting visual tokens $X_t^i$ describe local image regions; together with proprioception $q_t^i$, they provide evidence for local expert allocation and action prediction.

% Declare the double-column benchmark float on page 3 for placement on page 4.
% Shared source for the manuscript and the standalone experimental review.
\begin{figure*}[!htbp]
  \vspace*{5pt}
  \centering
  \includegraphics[width=0.94\textwidth]{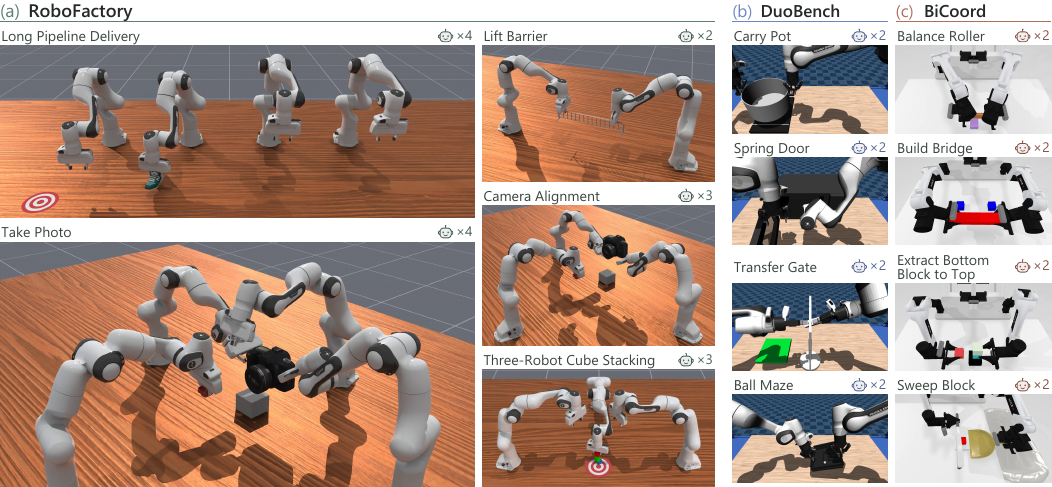}
  \caption{\textbf{Simulation benchmarks.} Thirteen tasks from RoboFactory~\cite{qin2025robofactory}, DuoBench~\cite{julg2026duobench}, and BiCoord~\cite{peng2026bicoord}; teams of two to four robots.}
  \label{fig:simulation-tasks}
  \vspace{-5pt}
\end{figure*}
\gdef\simulationtaskfigureplaced{}

\subsection{Learning Collaboration-Role Experts}
\label{sec:core-experts}

Local spatial evidence helps distinguish interaction situations, but the shared policy must still learn the complementary actions appropriate to them. In Fig.~\ref{fig:stereocore-overview}, CoRE learns collaboration-role experts with separately adjustable prediction paths, selects their contributions from local evidence, and uses demonstrated actions to supervise this selection. We omit robot index $i$.

\subsubsection{Constructing collaboration-role experts}
During a handover, maintaining a grasp and reaching to receive an object require different uses of visual evidence. To accommodate these complementary collaboration roles within one policy, CoRE uses an MoE structure~\cite{shazeer2017moe}, whose branches can learn different visual-to-action mappings.

CoRE equips each action-decoder layer with $E$ collaboration-role experts alongside shared cross-attention. Using a Transformer action-chunk regressor~\cite{zhao2023act}, it predicts $H$ future actions together. A learned \emph{action query} $u_k$ marks prediction position $k$. After self-attention in layer $\ell$, its feature is $h_{t,k}^{\ell}$. Expert $R_e^\ell(h_{t,k}^\ell,X_t)$ uses this feature to attend to visual tokens $X_t$ and returns a feature correction. Low-rank projections keep branches compact; a shared action head converts decoder features into actions. All robots share the expert pool; routing learns combinations conditioned on observations and action queries without predefined robot--expert or role--expert assignments.

\subsubsection{Choosing experts as collaboration unfolds}
An action chunk may span a transition from holding to releasing, so its prediction positions may require different expert combinations. A local router $g_\phi$, with parameters $\phi$, therefore considers both the current observation and the action query. It uses a multilayer perceptron (MLP) to score experts from the mean of $X_t$, an encoding of robot state $q_t$, and query $u_k$. Softmax gives normalized weights:
\begin{equation}
 p_{t,k}=\operatorname{softmax}\!\left(g_\phi(X_t,q_t,u_k)\right).
 \label{eq:local-router}
\end{equation}
We retain the top-$K$ experts, indexed by $\mathcal S_{t,k}$, and renormalize their weights as $\tilde p_{t,k,e}=p_{t,k,e}/\sum_{j\in\mathcal S_{t,k}}p_{t,k,j}$.

Their combined correction
\begin{equation}
 \Delta h_{t,k}^{\ell}=\sum_{e\in\mathcal S_{t,k}}
       \tilde p_{t,k,e}R_e^{\ell}(h_{t,k}^{\ell},X_t)
 \label{eq:expert-correction}
\end{equation}
is added to the shared cross-attention output before the feedforward block. This locally routed expert structure forms \emph{action-role cross-attention} (ARCA). All positions share the current observation, with position-specific routing weights shared across layers and recomputed for each new observation.

\subsubsection{Learning action--expert alignment}
Standard imitation backpropagates action error through the selected top-$K$ mixture, jointly updating action prediction and routing. Inspired by expert--router coupling~\cite{lv2026erc}, we introduce an \emph{action--expert alignment} (AEA) loss that complements this gradient signal with explicit action-space supervision from alternative routes. Under fixed decoder inputs, we activate each expert in turn across all layers and prediction positions, retaining shared computations and the action head. Full-chunk predictions yield per-position allocation targets, including for experts outside the current top-$K$. For demonstrated action $a_{t,k}^{*}$ and route prediction $\hat a_{t,k}^{(e)}$, we measure prediction error in normalized action space:
\begin{equation}
 \ell_{t,k,e}=\frac{1}{d_a}
       \left\|a_{t,k}^{*}-\hat a_{t,k}^{(e)}\right\|_2^2,
 \label{eq:expert-error}
\end{equation}
where $d_a$ is the action dimension. Comparing these errors at each position yields a supervision target favoring experts that better predict the required action:
\begin{equation}
 q_{t,k}^{\mathrm{cap}}=\operatorname{sg}\!\left[
       \operatorname{softmax}_{e}(-\ell_{t,k,e}/\tau_{t,k})\right].
 \label{eq:capability-target}
\end{equation}
Here $\tau_{t,k}$ is the standard deviation across expert errors, lower-bounded by $10^{-3}$; stop-gradient $\operatorname{sg}$ treats the target as fixed during differentiation. Kullback--Leibler (KL) divergence supervises the router probabilities before top-$K$ selection, averaged over the full chunk:
\begin{equation}
 \mathcal L_{\mathrm{AEA}}=\frac{1}{H}\sum_{k=1}^{H}
       D_{\mathrm{KL}}\!\left(q_{t,k}^{\mathrm{cap}}\,\|\,p_{t,k}\right).
 \label{eq:core-loss}
\end{equation}
With forced-route targets detached, the alignment loss supervises all pre-selection routing probabilities; imitation trains the selected action composition. All $H$ positions are supervised; padding repeats the final demonstrated action.

\subsection{Training and Deployment}

The conditional variational autoencoder (CVAE)~\cite{zhao2023act} encodes the demonstrated chunk into a training latent vector $z_t$. The same latent is used across expert-route comparisons and excluded from router inputs. We minimize
\begin{equation}
 \mathcal L=\mathcal L_{\mathrm{act}}
 +\lambda_{\mathrm{CVAE}}\mathcal L_{\mathrm{CVAE}}
 +\lambda_{\mathrm{AEA}}\mathcal L_{\mathrm{AEA}},
\end{equation}
where $\mathcal L_{\mathrm{act}}$ averages squared action error over valid steps and $\mathcal L_{\mathrm{CVAE}}$ regularizes the latent distribution toward a standard normal prior. We use $\lambda_{\mathrm{CVAE}}=10^{-3}$, $\lambda_{\mathrm{AEA}}=0.05$, four rank-32 experts, top-2 routing, and seven decoder layers.

Deployment sets $z_t=0$ and omits the expert-comparison branch. At each control step, every robot independently predicts a chunk of $H=100$ actions from its current local observation. Exponential temporal ensembling combines overlapping predictions for the current step with coefficient $0.01$. After executing one action, the robot observes the interaction again to update its prediction.
\endgroup
 % Expanded evidence-layout draft. Only measured entries carry numeric results.
% Figure and table slots marked PENDING are deliberately visible for review.
\section{Experiments}
\label{sec:experiments}

We organize the experiments around five research questions: \textbf{RQ1:} Does CoRE improve single-policy decentralized collaboration? \textbf{RQ2:} Which perception and decoder components contribute to performance? \textbf{RQ3:} Does expert allocation reflect collaborative action requirements? \textbf{RQ4:} How does collaboration change under partner perturbations? \textbf{RQ5:} Can CoRE support collaboration on physical robots?

\definecolor{expsage}{HTML}{607F70}
\definecolor{expclay}{HTML}{9E715D}
\definecolor{exppale}{HTML}{F5F4EF}
\newcommand{\expna}{\textcolor{expclay}{\textit{pending}}}
\newcommand{\exppending}[2]{%
  \begingroup\setlength{\fboxsep}{7pt}%
  \colorbox{exppale}{\parbox{\dimexpr\linewidth-14pt\relax}{%
  \small\raggedright\textcolor{expclay}{\textbf{PENDING EVIDENCE: #1}}\par\smallskip #2}}%
  \endgroup}

% Declare the double-column RQ1 table early so it shares the RQ1 page.
\begin{table*}[!t]
\vspace*{5pt}
\centering
\tabletitle{\textbf{Three-benchmark collaboration.} SSR/Avg.: percentages; SR: successes/100, no retries. Models: policy counts per benchmark. Superscripts: robot counts. Bold/underline: best/second-best (ties included). Reg.: RGB Regression; Central: joint observations.}
\label{tab:main-comparison}
\fontsize{8}{9}\selectfont
\setlength{\tabcolsep}{0.45pt}
\setlength{\arrayrulewidth}{0.3pt}
\renewcommand{\arraystretch}{1.0}
\begin{tabular*}{\textwidth}{@{\extracolsep{\fill}}l@{\hspace{2.5pt}}|@{\hspace{2.5pt}}c@{\hspace{2.5pt}}|@{\hspace{2.5pt}}*{4}{l}r@{\hspace{2.5pt}}|@{\hspace{2.5pt}}*{4}{l}r@{\hspace{2.5pt}}|@{\hspace{2.5pt}}*{5}{l}r@{}}
\toprule
\multirow{2}{*}{Method} & \multirow{2}{*}{Models}
& \multicolumn{5}{c@{\hspace{2.5pt}}|@{\hspace{2.5pt}}}{\textbf{DuoBench} (SSR $\uparrow$)}
& \multicolumn{5}{c@{\hspace{2.5pt}}|@{\hspace{2.5pt}}}{\textbf{BiCoord} (SR $\uparrow$)}
& \multicolumn{6}{c}{\textbf{RoboFactory} (SR $\uparrow$)} \\
& & Carry\textsuperscript{2} & Spring\textsuperscript{2} & Transfer\textsuperscript{2} & BallMaze\textsuperscript{2} & \textbf{Avg.}
& Balance\textsuperscript{2} & Bridge\textsuperscript{2} & Extract\textsuperscript{2} & Sweep\textsuperscript{2} & \textbf{Avg.}
& Lift\textsuperscript{2} & Camera\textsuperscript{3} & 3Stack\textsuperscript{3} & LPD\textsuperscript{4} & Photo\textsuperscript{4} & \textbf{Avg.} \\
\midrule
Central Reg. & 4/4/5
& \underline{66.3\%} & 35.7\% & \underline{13.8\%} & 51.5\% & 41.8\%
& \textbf{\srfrac{97}} & \textbf{\srfrac{78}} & \underline{\srfrac{69}} & \underline{\srfrac{33}} & \underline{69.3\%}
& \underline{\srfrac{97}} & \textbf{\srfrac{100}} & \srfrac{59} & \textbf{\srfrac{100}} & \textbf{\srfrac{96}} & \textbf{90.4\%} \\
Indep. Reg. & 8/8/16
& 34.1\% & 19.3\% & 10.8\% & 35.6\% & 25.0\%
& \srfrac{47} & \srfrac{22} & \srfrac{4} & \srfrac{11} & 21.0\%
& \srfrac{43} & \srfrac{19} & \srfrac{5} & \srfrac{0} & \srfrac{3} & 14.0\% \\
CLS-DP~\cite{park2026clsdp} & 8/8/16
& 63.3\% & \textbf{42.3\%} & 5.0\% & 52.0\% & 40.7\%
& \srfrac{22} & \srfrac{34} & \srfrac{0} & \srfrac{13} & 17.3\%
& \underline{\srfrac{97}} & \underline{\srfrac{96}} & \srfrac{53} & \srfrac{68} & \underline{\srfrac{55}} & 73.8\% \\
GauDP~\cite{wang2025gaudp} & 4/4/5
& 38.0\% & 32.3\% & 10.3\% & 29.0\% & 27.4\%
& \srfrac{21} & \srfrac{17} & \srfrac{0} & \srfrac{18} & 14.0\%
& \srfrac{72} & \srfrac{26} & \srfrac{0} & \srfrac{0} & \srfrac{3} & 20.2\% \\
CHORUS~\cite{doshi2026chorus} & 1/1/1
& 66.0\% & 32.3\% & \textbf{16.3\%} & \underline{59.5\%} & \underline{43.5\%}
& \srfrac{83} & \srfrac{16} & \srfrac{1} & \srfrac{6} & 26.5\%
& \textbf{\srfrac{99}} & \srfrac{89} & \underline{\srfrac{64}} & \srfrac{24} & \srfrac{16} & 58.4\% \\
\textbf{\method{} (Ours)} & 1/1/1
& \textbf{75.3\%} & \underline{37.7\%} & 13.3\% & \textbf{64.0\%} & \textbf{47.6\%}
& \underline{\srfrac{93}} & \underline{\srfrac{67}} & \textbf{\srfrac{70}} & \textbf{\srfrac{72}} & \textbf{75.5\%}
& \textbf{\srfrac{99}} & \textbf{\srfrac{100}} & \textbf{\srfrac{99}} & \underline{\srfrac{94}} & \srfrac{29} & \underline{84.2\%} \\
\bottomrule
\end{tabular*}
\vspace{-5pt}
\end{table*}

\subsection{Benchmarks and Evaluation Protocol}

\noindent\textbf{Benchmarks.}
In Figure~\ref{fig:simulation-tasks}, RoboFactory~\cite{qin2025robofactory}, DuoBench~\cite{julg2026duobench}, and BiCoord~\cite{peng2026bicoord} cover 13 tasks. RoboFactory's five tasks involve two to four robots: Long Pipeline Delivery (LPD) tests sequential transfer, Three-Robot Cube Stacking (ThreeStack) ordered placement, and Lift Barrier joint lifting. Camera Alignment and Take Photo require camera--object coordination; Take Photo adds a partner's approach to the camera. DuoBench's four tasks test coupled motion, asymmetric support, and constrained handovers; BiCoord's four test joint placement and object stabilization or reception. Each DuoBench/BiCoord arm executes independently from local observations.

\noindent\textbf{Compared methods.}
RGB Regression~\cite{zhao2023act} baselines vary policy sharing; Central jointly observes and controls the team. We also compare Stereo Regression with StereoPatch~\cite{zhou2026stereopatch}, GauDP~\cite{wang2025gaudp}, CHORUS~\cite{doshi2026chorus} with robot-identity prompts, and CLS-DP~\cite{park2026clsdp} with collaborative latents and task instructions.

\noindent\textbf{Training.}
Per benchmark, CoRE pools all task and robot demonstrations to train one policy. RoboFactory uses 120,000 AdamW updates, batch size 40, hierarchical task balancing, and policy/router learning rates $2\times10^{-4}/3\times10^{-4}$. Alignment targets use all routes on one sample every four updates; imitation uses the full batch every update.

\noindent\textbf{Evaluation metrics.}
Stage success ratio (SSR) averages the ratio of highest stage reached to total stages; success rate (SR) measures full-task completion. Methods share 100 seeds per task, one rollout each; tasks have equal weight.

\noindent\textbf{Physical setup.}
Two Piper arms operate as independent robots running copies of one shared policy, using AR0234 stereo sensing~\cite{zhou2026stereopatch}. Each robot uses top-view RGB-D, its wrist RGB, and proprioception to independently predict its own actions. Five tasks have 30 demonstrations each; training uses 80,000 updates, batch size 32, and $640\times320$ RGB ($20\times40$ patches).

% Standalone reviews include the figure here; main.tex declares it earlier.
\ifdefined\simulationtaskfigureplaced\else
  % Shared source for the manuscript and the standalone experimental review.
\begin{figure*}[!htbp]
  \vspace*{5pt}
  \centering
  \includegraphics[width=0.94\textwidth]{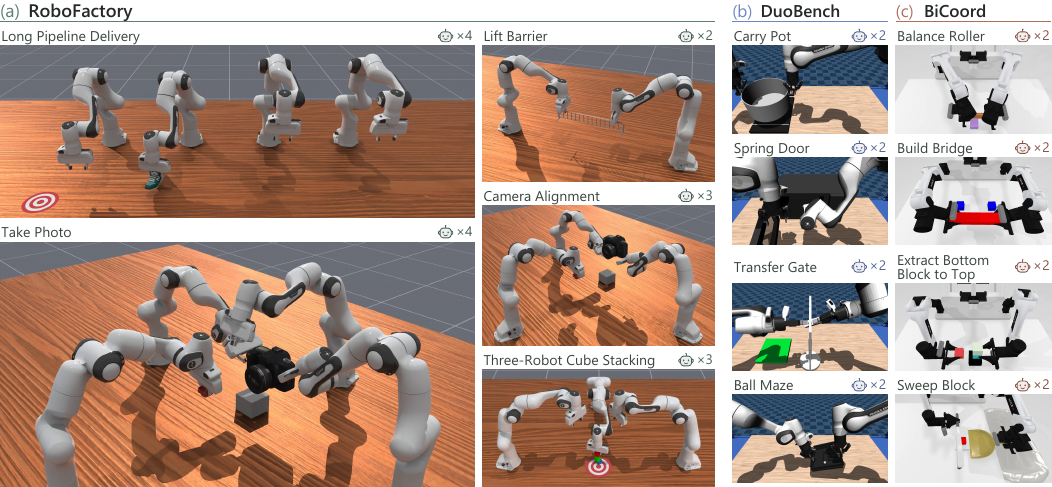}
  \caption{\textbf{Simulation benchmarks.} Thirteen tasks from RoboFactory~\cite{qin2025robofactory}, DuoBench~\cite{julg2026duobench}, and BiCoord~\cite{peng2026bicoord}; teams of two to four robots.}
  \label{fig:simulation-tasks}
  \vspace{-5pt}
\end{figure*}
\gdef\simulationtaskfigureplaced{}

\fi

\setlength{\textfloatsep}{6pt plus 2pt minus 1pt}

\subsection{RQ1: Single-Policy Decentralized Collaboration}
\label{sec:rq1}

\noindent\textbf{Collaboration with one shared policy.}
In Table~\ref{tab:main-comparison}, one CoRE policy learns all selected tasks and robot behaviors within each benchmark. Central regression jointly predicts team actions from joint observations, achieving 90.4\% success on RoboFactory. Under decentralized execution, CoRE reaches 84.2\% with identical policy copies acting independently from local observations. It achieves the highest mean performance among evaluated decentralized methods on all three benchmarks, supporting the effectiveness of learning complementary team behaviors within one shared policy.

\begin{figure}[!t]
\centering
\includegraphics[width=\columnwidth]{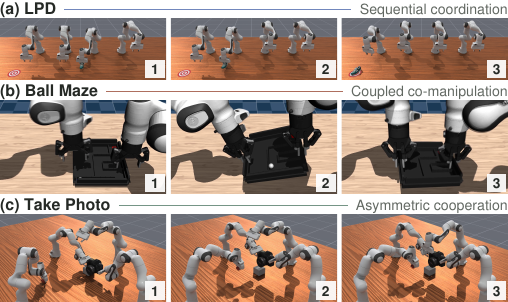}
\caption{Representative collaboration requirements illustrated through task sequences, progressing from left to right.}
\label{fig:collaboration-sequences}
\iflatexml\end{figure}\begin{table}\fi
\par\vspace{4pt}
\begingroup
\centering
\tabletitle{\textbf{Demonstration sharing.} RoboFactory; metrics as in Table~\ref{tab:main-comparison}. Data denotes pooling scope; Ours adds depth.}
\label{tab:sharing-scope}
\fontsize{8}{9}\selectfont
\setlength{\tabcolsep}{0.35pt}
\setlength{\arrayrulewidth}{0.3pt}
\renewcommand{\arraystretch}{1.0}
\begin{tabular*}{\columnwidth}{@{\extracolsep{\fill}}l@{\hspace{3pt}}lc*{5}{r}r@{}}
\toprule
Method & Data & Models & Lift & Camera & 3Stack & LPD & Photo & \textbf{Avg.} \\
\midrule
RGB Regression & Task/robot & 16 & \srfrac{43} & \srfrac{19} & \srfrac{5} & \srfrac{0} & \srfrac{3} & 14.0\% \\
RGB Regression & Per task & 5 & \srfrac{86} & \srfrac{94} & \srfrac{26} & \srfrac{48} & \srfrac{14} & 53.6\% \\
RGB Regression & All tasks & 1 & \srfrac{87} & \underline{\srfrac{98}} & \srfrac{3} & \srfrac{0} & \srfrac{14} & 40.4\% \\
CoRE (RGB) & All tasks & 1 & \underline{\srfrac{89}} & \textbf{\srfrac{100}} & \underline{\srfrac{71}} & \underline{\srfrac{86}} & \underline{\srfrac{16}} & \underline{72.4\%} \\
\midrule
\textbf{CoRE (Ours)} & All tasks & 1 & \textbf{\srfrac{99}} & \textbf{\srfrac{100}} & \textbf{\srfrac{99}} & \textbf{\srfrac{94}} & \textbf{\srfrac{29}} & \textbf{84.2\%} \\
\bottomrule
\end{tabular*}
\endgroup
\iflatexml\end{table}\else\end{figure}\fi

% Queue the full-width RQ3 figure on page 5 for placement at the top of page 6.
\begin{figure*}[!t]
\vspace*{5pt}
\centering
\includegraphics[width=\textwidth]{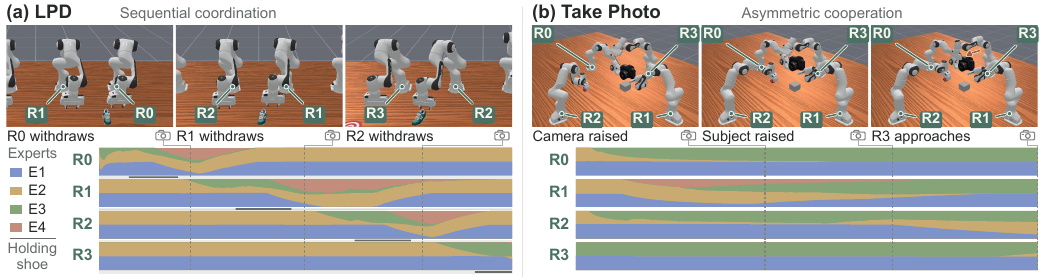}
\caption{Shared expert use during collaboration. (a) LPD: sequential object transfers, with robots numbered in transfer order. (b) Take Photo: asymmetric cooperation. Colors show action-history-weighted expert summaries; gray bars indicate grasp intervals. Snapshots align with dashed time guides.}
\label{fig:routing-mechanism}
\vspace{-5pt}
\end{figure*}

\noindent\textbf{Learning from team demonstrations.}
In Figure~\ref{fig:collaboration-sequences}, LPD illustrates sequential object transfer, Ball Maze coupled manipulation, and Take Photo asymmetric cooperation. These tasks require different forms of coordination: passing an object onward as partners become ready, moving a shared object together, or maintaining camera--subject alignment while another robot approaches the camera. Table~\ref{tab:sharing-scope} examines whether pooling team demonstrations helps one policy learn these interdependent behaviors. Independent regression has limited success across the five tasks. Pooling demonstrations within each task improves all five, with particularly large gains on Camera and LPD. Team experience thus benefits both relative-pose coordination and sequential action responsibilities. Extending pooling to all tasks largely removes the gains on LPD and ThreeStack, while Lift and Camera remain strong. This contrast highlights the challenge of accommodating diverse collaborative behaviors within one policy. With RGB inputs and the cross-task demonstration pool held fixed, CoRE raises mean success from 40.4\% to 72.4\%. This supports learning expert organization and selection to accommodate diverse collaborative behaviors within one policy. Adding appearance--geometry perception further improves performance, most strongly on ThreeStack, consistent with the spatial demands of coordinated placement. Even without depth, one CoRE policy outperforms the five task-specific regression policies on every task, supporting its ability to learn complementary team behaviors across tasks. Take Photo remains less successful, reflecting the difficulty of maintaining camera--subject alignment while coordinating a partner's approach.

\subsection{RQ2: Component Contributions and Routing Sparsity}
\label{sec:rq2}
\setlength{\textfloatsep}{6pt plus 2pt minus 1pt}

\noindent\textbf{Learning behaviors and expert selection.}
Table~\ref{tab:module-ablation} distinguishes spatial perception, expert prediction paths, and expert-selection supervision. Spatial perception improves ThreeStack, and expert architectures extend this gain, while LPD remains weak without alignment supervision. With spatial inputs and cross-attention experts held fixed, adding action--expert alignment in full CoRE raises LPD success from 15\% to 94\%. This contrast highlights the role of demonstrated-action supervision in learning which experts to select from local observations, alongside learning separately adaptable prediction paths. CoRE (RGB) already performs strongly on both sequential tasks; adding spatial perception in full CoRE further improves ThreeStack. These results support the complementary contributions of spatial representations and action-guided expert selection to collaborative learning.

\begin{table}[!t]
\centering
\tabletitle{\textbf{Module ablations.} RoboFactory with one shared policy. Task scores, averages, and ranking marks follow Table~\ref{tab:main-comparison}.}
\label{tab:module-ablation}
\fontsize{8}{9}\selectfont
\setlength{\tabcolsep}{1.2pt}
\renewcommand{\arraystretch}{1.0}
\begin{tabular*}{\columnwidth}{@{\extracolsep{\fill}}l*{5}{l}r@{}}
\toprule
Variant & Lift & Camera & 3Stack & LPD & Photo & Avg. \\
\midrule
RGB Regression & \srfrac{87} & \srfrac{98} & \srfrac{3} & \srfrac{0} & \srfrac{14} & 40.4\% \\
Stereo Regression & \srfrac{92} & \textbf{\srfrac{100}} & \srfrac{42} & \srfrac{20} & \textbf{\srfrac{29}} & 56.6\% \\
Stereo + MoE & \textbf{\srfrac{100}} & \underline{\srfrac{99}} & \srfrac{68} & \srfrac{17} & \underline{\srfrac{28}} & 62.4\% \\
Stereo + MoE + ARCA & \srfrac{96} & \textbf{\srfrac{100}} & \underline{\srfrac{81}} & \srfrac{15} & \textbf{\srfrac{29}} & 64.2\% \\
CoRE (RGB) & \srfrac{89} & \textbf{\srfrac{100}} & \srfrac{71} & \underline{\srfrac{86}} & \srfrac{16} & \underline{72.4\%} \\
\textbf{CoRE (Ours)} & \underline{\srfrac{99}} & \textbf{\srfrac{100}} & \textbf{\srfrac{99}} & \textbf{\srfrac{94}} & \textbf{\srfrac{29}} & \textbf{84.2\%} \\
\bottomrule
\end{tabular*}
\end{table}

\begin{table}[!t]
\centering
\tabletitle{\textbf{Expert count and routing sparsity.} Settings and metrics as in Table~\ref{tab:module-ablation}; $E$ experts, top-$K$ routing; *: our default.}
\label{tab:expert-sweep}
\fontsize{8}{9}\selectfont
\setlength{\tabcolsep}{1.2pt}
\renewcommand{\arraystretch}{1.0}
\begin{tabular*}{\columnwidth}{@{\extracolsep{\fill}}ll*{5}{l}r@{}}
\toprule
$E$ & $K$ & Lift & Camera & 3Stack & LPD & Photo & Avg. \\
\midrule
2 & 2 & \srfrac{96} & \textbf{\srfrac{100}} & \textbf{\srfrac{99}} & \srfrac{65} & \textbf{\srfrac{41}} & 80.2\% \\
4 & 1 & \srfrac{87} & \underline{\srfrac{96}} & \srfrac{82} & \srfrac{33} & \srfrac{15} & 62.6\% \\
4\textsuperscript{*} & 2 & \underline{\srfrac{99}} & \textbf{\srfrac{100}} & \textbf{\srfrac{99}} & \underline{\srfrac{94}} & \srfrac{29} & \textbf{84.2\%} \\
4 & 4 & \textbf{\srfrac{100}} & \textbf{\srfrac{100}} & \srfrac{59} & \srfrac{4} & \underline{\srfrac{31}} & 58.8\% \\
8 & 2 & \underline{\srfrac{99}} & \textbf{\srfrac{100}} & \underline{\srfrac{98}} & \textbf{\srfrac{97}} & \srfrac{23} & \underline{83.4\%} \\
\bottomrule
\end{tabular*}
\end{table}

\noindent\textbf{Expert capacity and routing sparsity.}
In Table~\ref{tab:expert-sweep}, $E$ determines how many experts are available, while $K$ determines how many contribute to each predicted action. At $K=2$, increasing $E$ from two to four improves LPD but reduces Take Photo performance; eight experts offer no further mean improvement. At $E=4$, top-2 achieves higher mean success than top-1 or activating all experts, with dense routing particularly degrading LPD. Among the tested configurations, maintaining multiple experts with selective composition benefits collaboration more than increasing simultaneous activation. We therefore use $(E,K)=(4,2)$.

\subsection{RQ3: Shared Expert Use during Collaboration}
\label{sec:rq3}

\noindent\textbf{Sequential reuse across robots.}
In Figure~\ref{fig:routing-mechanism}(a), R0, R1, and R2 successively release the object and withdraw during LPD. Transient increases in E4 and reductions in E1 recur near these transitions. This pattern follows the same action requirement across robots, while each robot changes its mixture through the transfer stages. These trajectories illustrate reuse of the shared expert pool across robots and adjustment of expert contributions as their action responsibilities change.

\noindent\textbf{Concurrent complementary behavior.}
In Figure~\ref{fig:routing-mechanism}(b), Take Photo requires robots to maintain the camera--subject arrangement while a partner approaches the camera. Near completion, R0 and R1 remain nearly stationary as R3 continues approaching. All three use similar E1/E3 mixtures, while R2 retains a larger E2 contribution as it holds the subject. Local observations and action queries condition the shared experts' outputs, enabling each robot to use this common pool for its current collaborative action.

\subsection{RQ4: Robustness of Decentralized Collaboration}
\label{sec:rq4}

\begingroup
\begin{figure}[!htbp]
\centering
\includegraphics[width=\columnwidth]{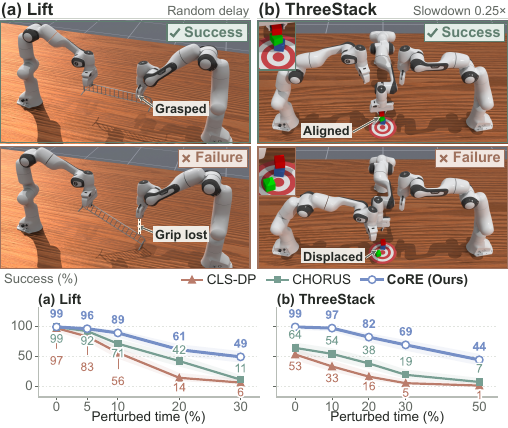}
\caption{\textbf{Collaboration robustness.} SR versus perturbed time (\%) for random delay (Lift) and slowdown (ThreeStack); bands: 95\% Wilson intervals.}
\label{fig:partner-behavior}
\end{figure}
\endgroup

% Queue the real-world result panel on page 6 for the top of page 7.
\begin{figure*}[!t]
\vspace*{5pt}
\centering
\includegraphics[width=\textwidth,trim=0bp 3.32bp 0bp 1.72bp,clip]{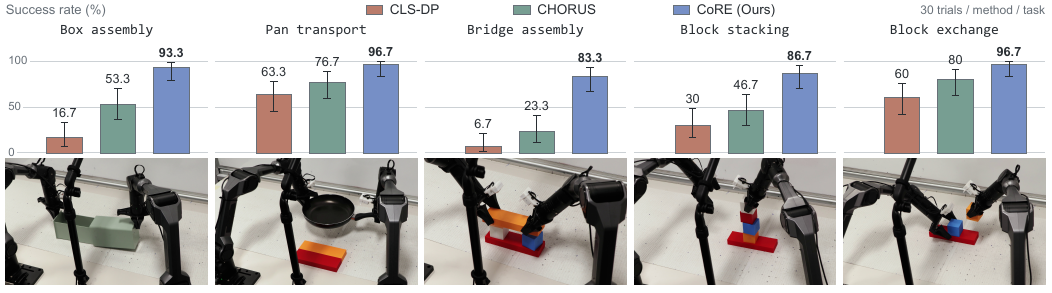}
\caption{\textbf{Real-world collaborative manipulation.} Top: success rates over 30 trials per method and task; error bars denote 95\% Wilson intervals. Bottom: representative task demonstrations.}
\label{fig:physical-results}
\vspace{-5pt}
\end{figure*}

\noindent\textbf{Perturbing partner execution.}
In Figure~\ref{fig:partner-behavior}, we introduce random partner delays during Lift and randomly slow one ThreeStack robot to $\speedratio{0.25}$ speed, inspired by SAI's partner-delay evaluation~\cite{chen2026sai}. The horizontal axis gives the percentage of time affected; each point uses 100 trials. Lift requires maintaining a shared grasp during joint motion, whereas ThreeStack requires successive placements on a stable stack.

\noindent\textbf{Robustness across collaboration requirements.}
In Figure~\ref{fig:partner-behavior}, CoRE achieves the highest success at every nonzero perturbation level. On Lift, CoRE and CHORUS start from identical nominal success, but CoRE degrades less under delay. On ThreeStack, CoRE also retains a larger fraction of its nominal success as slowdown increases. This advantage is consistent with action--expert alignment training the local router to select experts according to observed collaboration requirements. The relevant demands differ across tasks: maintaining a shared grasp during delayed joint motion in Lift, and coordinating subsequent placements with a slowed partner's progress in ThreeStack. The results show greater tolerance to partner timing disruptions in both coupled and sequential collaboration.

\subsection{RQ5: Collaboration and Robustness on Physical Robots}
\label{sec:rq5}

Figure~\ref{fig:physical-results} summarizes five physical tasks performed by two independently acting Piper robots. Box assembly combines reorientation and joining of two parts; pan transport requires moving a shared object onto supports. Bridge assembly and block stacking involve coordinated placement, while block exchange reverses the blocks' stacking order through coordinated rearrangement. Together, they test coupled, asymmetric, and sequential collaboration.

Across 30 trials per method and task, CoRE achieves 91.3\% success, versus 56.0\% for CHORUS and 35.3\% for CLS-DP, and leads on all five tasks. Bridge assembly improves from CHORUS's 7/30 to 25/30, demonstrating an advantage in coordinated placement. The five-task results show that a shared policy supports relative-pose coordination, shared-object transport, and collaborative rearrangement.

Figure~\ref{fig:physical-box}(a) shows the transition from coupled flipping to asymmetric joining in box assembly. The two robots first coordinate part reorientation, then follow different joining trajectories while maintaining mating alignment. Their action responsibilities must therefore change with the interaction stage. CoRE deploys independent copies of the shared policy, each selecting expert contributions from local observations to predict its robot's actions.

Figure~\ref{fig:physical-box}(b) highlights sequential collaboration, as in the simulated ThreeStack task. The arms take turns placing blocks, and each completed placement establishes the support surface for the next. The placing robot's action thus depends on the configuration left by its partner. Whereas box assembly changes collaboration mode within a task, stacking repeatedly transfers the placement responsibility between arms. Both require action predictions to reflect the evolving interaction state, with different simultaneous motions in joining and ordered contributions in stacking.

\begingroup
\begin{figure}[!htbp]
\centering
\setlength{\abovecaptionskip}{1.5pt}
\includegraphics[width=\columnwidth]{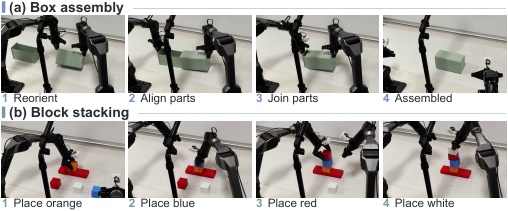}
\caption{\textbf{Physical collaboration.} Coupled flipping followed by asymmetric joining in box assembly (top), and sequential block placement (bottom).}
\label{fig:physical-box}
\end{figure}
\endgroup

\begingroup
\begin{figure}[!htbp]
\centering
\includegraphics[width=\columnwidth,trim=0bp 0.16bp 0bp 1.00bp,clip]{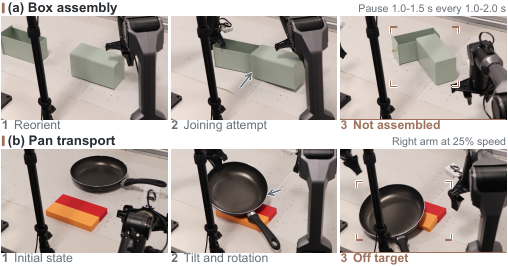}
\caption{\textbf{Real-world collaboration robustness.} Failure sequences under random partner pauses (Box, top) and $\speedratio{0.25}$ partner speed (Pan, bottom).}
\label{fig:physical-failures}
\iflatexml\end{figure}\begin{figure}\fi
\par\vspace{4pt}
\centering
\includegraphics[width=\columnwidth]{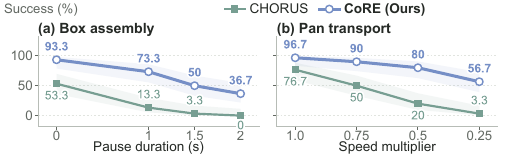}
\caption{\textbf{Physical collaboration robustness.} SR under periodic pauses (Box) and reduced speed (Pan); 30 trials per point, with 95\% Wilson intervals.}
\label{fig:physical-robustness}
\end{figure}
\endgroup

The failure sequences in Figure~\ref{fig:physical-failures} illustrate how partner timing affects physical coordination. During box assembly, pauses interrupt joining and the halves remain misaligned. During pan transport, slowing the right arm to $\speedratio{0.25}$ tilts and rotates the pan, leaving it offset from its supports. Both failures disrupt the required object alignment.

In Figure~\ref{fig:physical-robustness}, we quantify robustness over 30 trials per condition. In Box, the right robot pauses for 1.0, 1.5, or 2.0\,s after every 2.0\,s of normal execution. In Pan, execution speed is reduced to $\speedratio{0.75}$, $\speedratio{0.5}$, or $\speedratio{0.25}$. CoRE outperforms CHORUS at every perturbation level. With 1.0\,s pauses, CoRE completes 22/30 trials versus 4/30 for CHORUS; at $\speedratio{0.25}$ speed, it completes 17/30 versus 1/30. Intermittent pauses challenge the alignment of joining actions, whereas sustained slowdown challenges shared-object transport. Success decreases as pauses lengthen or partner motion slows, but CoRE retains more of its nominal performance across both tasks. These results demonstrate greater tolerance to timing disruptions in asymmetric assembly and coupled transport.

\FloatBarrier

 \section{Conclusion}

CoRE learns single-policy decentralized collaboration from pooled multi-task, multi-robot demonstrations. Collaboration-role experts provide adaptable paths for complementary behaviors; action--expert alignment supervises local expert selection using forced-route prediction errors against demonstrations, without role labels. Demonstration-sharing comparisons highlight the importance of learning how to organize and select team behaviors as training expands across robots and tasks. Across simulation benchmarks and physical manipulation tasks, identical policy copies support coupled, asymmetric, and sequential collaboration with improved tolerance to partner timing disruptions. Together, these results show how demonstrated actions can guide both behavior learning and expert selection, enabling robots to coordinate through a shared policy conditioned on their own observations as collaborative responsibilities evolve.

\FloatBarrier

\bibliographystyle{IEEEtran}
\bibliography{references}

@IEEEtranBSTCTL{CoREBSTcontrol,
  CTLuse_forced_etal={yes},
  CTLmax_names_forced_etal={6},
  CTLnames_show_etal={1}
}

@inproceedings{lowe2017maddpg,
  author={Lowe, Ryan and Wu, Yi and Tamar, Aviv and Harb, Jean and Abbeel, Pieter and Mordatch, Igor},
  title={Multi-Agent Actor-Critic for Mixed Cooperative-Competitive Environments},
  booktitle={Adv. Neural Inf. Process. Syst.},
  volume={30},
  year={2017},
  sourceurl={https://proceedings.neurips.cc/paper_files/paper/2017/hash/68a9750337a418a86fe06c1991a1d64c-Abstract.html},
}

@article{zhou2026stereopatch,
  author={Zhou, Yanan and Qian, Zhaoyan and Zhao, James and Zhi, Weiming},
  title={{StereoPatch}: Patch-Aligned {RGB}-Depth Fusion for Spatial Perception in Robot Manipulation},
  journal={arXiv preprint arXiv:2609.15509},
  doi={10.48550/arXiv.2609.15509},
  sourceurl={https://arxiv.org/abs/2609.15509},
  year={2026}
}

@inproceedings{chi2023diffusionpolicy,
  author={Chi, Cheng and Feng, Siyuan and Du, Yilun and Xu, Zhenjia and Cousineau, Eric and Burchfiel, Benjamin and Song, Shuran},
  title={{Diffusion Policy}: Visuomotor Policy Learning via Action Diffusion},
  booktitle={Proc. Robotics: Science and Systems (RSS)},
  doi={10.15607/RSS.2023.XIX.026},
  sourceurl={https://www.roboticsproceedings.org/rss19/p026.html},
  year={2023}
}

@inproceedings{black2024pi0,
  author={Black, Kevin and Brown, Noah and Driess, Danny and Esmail, Adnan and Equi, Michael Robert and Finn, Chelsea and Fusai, Niccolo and Groom, Lachy and Hausman, Karol and Ichter, Brian and Jakubczak, Szymon and Jones, Tim and Ke, Liyiming and Levine, Sergey and Li-Bell, Adrian and Mothukuri, Mohith and Nair, Suraj and Pertsch, Karl and Shi, Lucy Xiaoyang and Smith, Laura and Tanner, James and Vuong, Quan and Walling, Anna and Wang, Haohuan and Zhilinsky, Ury},
  title={{\(\pi_0\)}: A Vision-Language-Action Flow Model for General Robot Control},
  booktitle={Proc. Robotics: Science and Systems (RSS)},
  doi={10.15607/RSS.2025.XXI.010},
  sourceurl={https://www.roboticsproceedings.org/rss21/p010.html},
  year={2025}
}

@article{caccavale2008impedance,
  author={Caccavale, Fabrizio and Chiacchio, Pasquale and Marino, Alessandro and Villani, Luigi},
  title={Six-{DOF} Impedance Control of Dual-Arm Cooperative Manipulators},
  journal={IEEE/ASME Trans. Mechatronics},
  volume={13},
  number={5},
  pages={576--586},
  doi={10.1109/TMECH.2008.2002816},
  year={2008}
}

@inproceedings{tsiamis2015implicit,
  author={Tsiamis, Anastasios and Verginis, Christos K. and Bechlioulis, Charalampos P. and Kyriakopoulos, Kostas J.},
  title={Cooperative Manipulation Exploiting Only Implicit Communication},
  booktitle={Proc. IEEE/RSJ Int. Conf. Intelligent Robots and Systems (IROS)},
  pages={864--869},
  doi={10.1109/IROS.2015.7353473},
  year={2015}
}

@inproceedings{qin2025robofactory,
  author={Qin, Yiran and Kang, Li and Song, Xiufeng and Yin, Zhenfei and Liu, Xiaohong and Liu, Xihui and Zhang, Ruimao and Bai, Lei},
  title={{RoboFactory}: Exploring Embodied Agent Collaboration with Compositional Constraints},
  booktitle={Proc. IEEE/CVF Int. Conf. Computer Vision (ICCV)},
  pages={10075--10085},
  year={2025}
}

@inproceedings{wang2025gaudp,
  author={Wang, Ziye and Kang, Li and Qin, Yiran and Ma, Jiahua and Peng, Zhanglin and Bai, Lei and Zhang, Ruimao},
  title={{GauDP}: Reinventing Multi-Agent Collaboration through {Gaussian}-Image Synergy in Diffusion Policies},
  booktitle={Adv. Neural Inf. Process. Syst.},
  volume={38},
  pages={5620--5639},
  doi={10.52202/085713-0198},
  sourceurl={https://papers.neurips.cc/paper_files/paper/2025/hash/0895de150660b6c2aef9b0f098cb1dcd-Abstract-Conference.html},
  year={2025}
}

@inproceedings{he2025latenttom,
  author={He, Chengyang and Camps, Gadiel Mark Sznaier and Liu, Xu and Schwager, Mac and Sartoretti, Guillaume Adrien},
  title={Latent Theory of Mind: A Decentralized Diffusion Architecture for Cooperative Manipulation},
  booktitle={Proc. Conf. Robot Learning (CoRL)},
  series={PMLR},
  volume={305},
  pages={392--405},
  year={2025}
}

@article{park2026clsdp,
  author={Park, Chanyoung and Yoon, Minsung and Jeong, Andrew and Yoon, Sung-eui},
  title={Distilling Collaborative Dynamics into Latent Space for Implicit Coordination in Decentralized Multi-Agent Manipulation},
  journal={arXiv preprint arXiv:2606.22982},
  note={Accepted to IROS},
  year={2026}
}

@article{doshi2026chorus,
  title={{CHORUS}: Decentralized Multi-Embodiment Collaboration with One {VLA} Policy},
  author={Doshi, Ria and Gao, Tian and Chen, Annie and Finn, Chelsea and Bohg, Jeannette},
  journal={arXiv preprint arXiv:2606.12352},
  year={2026}
}

@article{julg2026duobench,
  author={J{\"u}lg, Tobias and Bien, Seongjin and Hilber, Simon and Blei, Yannik and Krack, Pierre and Li, Maximilian and Parusel, Sven and Lioutikov, Rudolf and Walter, Florian and Burgard, Wolfram},
  title={{DuoBench}: A Reproducible Benchmark for Bimanual Manipulation in Simulation and the Real World},
  journal={arXiv preprint arXiv:2606.11901},
  year={2026}
}

@article{peng2026bicoord,
  author={Peng, Xingyu and Gao, Chen and Jin, Liankai and Li, Annan and Liu, Si},
  title={{BiCoord}: A Bimanual Manipulation Benchmark towards Long-Horizon Spatial--Temporal Coordination},
  journal={arXiv preprint arXiv:2604.05831},
  year={2026}
}

@inproceedings{howell2023capabilities,
  author={Howell, Pierce and Rudolph, Max and Torbati, Reza Joseph and Fu, Kevin and Ravichandar, Harish},
  title={Generalization of Heterogeneous Multi-Robot Policies via Awareness and Communication of Capabilities},
  booktitle={Proc. Conf. Robot Learning (CoRL)},
  series={PMLR},
  volume={229},
  pages={2772--2790},
  year={2023}
}

@inproceedings{shazeer2017moe,
  author={Shazeer, Noam and Mirhoseini, Azalia and Maziarz, Krzysztof and Davis, Andy and Le, Quoc V. and Hinton, Geoffrey E. and Dean, Jeff},
  title={Outrageously Large Neural Networks: The Sparsely-Gated Mixture-of-Experts Layer},
  booktitle={Proc. Int. Conf. Learning Representations (ICLR)},
  year={2017}
}

@article{jacobs1991adaptive,
  author={Jacobs, Robert A. and Jordan, Michael I. and Nowlan, Steven J. and Hinton, Geoffrey E.},
  title={Adaptive Mixtures of Local Experts},
  journal={Neural Comput.},
  volume={3},
  number={1},
  pages={79--87},
  doi={10.1162/neco.1991.3.1.79},
  year={1991}
}

@inproceedings{kada2026tgr,
  author={Kada, Masahiro and Yoshihashi, Ryota and Ikehata, Satoshi and Kawakami, Rei and Sato, Ikuro},
  title={Teacher-Guided Routing for Sparse Vision Mixture-of-Experts},
  booktitle={Proc. IEEE/CVF Conf. Computer Vision and Pattern Recognition (CVPR)},
  pages={6198--6208},
  year={2026}
}

@article{fedus2022switch,
  author={Fedus, William and Zoph, Barret and Shazeer, Noam},
  title={{Switch Transformers}: Scaling to Trillion Parameter Models with Simple and Efficient Sparsity},
  journal={J. Mach. Learn. Res.},
  volume={23},
  number={120},
  pages={1--39},
  year={2022}
}

@inproceedings{zhou2022expertchoice,
  author={Zhou, Yanqi and Lei, Tao and Liu, Hanxiao and Du, Nan and Huang, Yanping and Zhao, Vincent and Dai, Andrew M. and Chen, Zhifeng and Le, Quoc V. and Laudon, James},
  title={Mixture-of-Experts with Expert Choice Routing},
  booktitle={Adv. Neural Inf. Process. Syst.},
  volume={35},
  pages={7103--7114},
  year={2022}
}

@inproceedings{wang2025sparsedp,
  author={Wang, Yixiao and Zhang, Yifei and Huo, Mingxiao and Tian, Thomas and Zhang, Xiang and Xie, Yichen and Xu, Chenfeng and Ji, Pengliang and Zhan, Wei and Ding, Mingyu and Tomizuka, Masayoshi},
  title={{Sparse Diffusion Policy}: A Sparse, Reusable, and Flexible Policy for Robot Learning},
  booktitle={Proc. Conf. Robot Learning (CoRL)},
  series={PMLR},
  volume={270},
  pages={649--665},
  year={2025}
}

@article{rodriguez2026larmoe,
  author={Rodriguez, Ariel and Li, Chenpan and Mazza, Lorenzo and Younis, Rayan and Hellig, Ortrun and Bodenstedt, Sebastian and Wagner, Martin and Speidel, Stefanie},
  title={{LAR-MoE}: Latent-Aligned Routing for Mixture of Experts in Robotic Imitation Learning},
  journal={arXiv preprint arXiv:2603.08476},
  year={2026}
}

@article{huang2026corde,
  author={Huang, Haidong and Zhao, Xixin and Zhou, Yaohua and Song, Jiayu and Zhang, Jiayi and Ma, Jun and Zhu, Haiyue and Li, Xiaocong},
  title={{CoRDE}: Concept-Prior Routed Diffusion Experts for Structural Generalization in Robot Manipulation},
  journal={arXiv preprint arXiv:2606.21935},
  note={Accepted to IROS},
  year={2026}
}

@article{yoon2026misrouted,
  author={Yoon, Youngsik and Wang, Siwei and Chen, Wei and Ok, Jungseul},
  title={When Are Experts Misrouted? {Counterfactual} Routing Analysis in Mixture-of-Experts Language Models},
  journal={arXiv preprint arXiv:2605.07260},
  year={2026}
}

@inproceedings{lv2026erc,
  author={Lv, Ang and Ma, Jin and Ma, Yiyuan and Qiao, Siyuan},
  title={Coupling Experts and Routers in Mixture-of-Experts via an Auxiliary Loss},
  booktitle={Proc. Int. Conf. Learning Representations (ICLR)},
  year={2026}
}

@inproceedings{zhao2023act,
  author={Zhao, Tony Z. and Kumar, Vikash and Levine, Sergey and Finn, Chelsea},
  title={Learning Fine-Grained Bimanual Manipulation with Low-Cost Hardware},
  booktitle={Proc. Robotics: Science and Systems (RSS)},
  doi={10.15607/RSS.2023.XIX.016},
  year={2023}
}

@article{simeoni2025dinov3,
  title={{DINOv3}},
  author={Oriane Sim{\'e}oni and Huy V. Vo and Maximilian Seitzer and Federico Baldassarre and Maxime Oquab and Cijo Jose and Vasil Khalidov and Marc Szafraniec and Seung Eun Yi and Michael Ramamonjisoa and Francisco Massa and Daniel Haziza and Luca Wehrstedt and Jianyuan Wang and Timoth{\'e}e Darcet and Th{\'e}o Moutakanni and Leonel Sentana and Claire Roberts and Andrea Vedaldi and Jamie Tolan and John Brandt and Camille Couprie and Julien Mairal and Herve Jegou and Patrick Labatut and Piotr Bojanowski},
  journal={Trans. Mach. Learn. Res.},
  issn={2835-8856},
  sourceurl={https://openreview.net/forum?id=2NlGyqNjns},
  year={2026}
}

@article{patel2026defm,
  author={Patel, Manthan and Frey, Jonas and Mittal, Mayank and Yang, Fan and Hansson, Alexander and Bar, Amir and Cadena, Cesar and Hutter, Marco},
  title={{DeFM}: Learning Foundation Representations from Depth for Robotics},
  journal={arXiv preprint arXiv:2601.18923},
  year={2026}
}

@article{chen2026sai,
  author={Chen, Yincong and Qiu, Ranpeng and Li, Zihao and Zhou, Yanan and Ren, Guoqiang and Zhi, Weiming},
  title={Robots that Collaborate: Sequential Asymmetric Imitation for Learning Coupled Robot Policies},
  journal={arXiv preprint arXiv:2606.16490},
  note={Accepted to CoRL},
  year={2026}
}

@article{zhou2026patch,
  author={Zhou, Yanan and Qiu, Ranpeng and Chen, Yincong and Cui, Jiajie and Zhi, Weiming},
  title={{PATCH}: Action-Chunk-Conditioned Latent Patch Innovation Monitoring for Robot Manipulation},
  journal={arXiv preprint arXiv:2606.16690},
  year={2026}
}

@article{zhao2026trimanpolicy,
  author={Zhao, James and Ba, Mingyuan and Zhi, Weiming},
  title={Tri-Manual Visuomotor Imitation Learning of Robot Policies},
  journal={arXiv preprint arXiv:2607.25731},
  year={2026}
}

@article{dong2025mimicd,
  author={Dong, Dayi and Bhatt, Maulik and Choi, Seoyeon and Mehr, Negar},
  title={{MIMIC-D}: Multi-modal Imitation for Multi-agent Coordination with Decentralized Diffusion Policies},
  journal={arXiv preprint arXiv:2509.14159},
  note={Accepted to ICRA 2026},
  year={2025}
}

\end{document}